\documentclass[letterpaper, 10 pt, conference]{ieeeconf}  

\IEEEoverridecommandlockouts                              
\usepackage{graphics} 
\usepackage{epsfig} 
\usepackage{mathptmx} 
\usepackage{times} 
\usepackage{amsmath} 
\usepackage{amssymb}  
\usepackage{xspace}

\title{\LARGE \bf
Deep auxiliary learning for visual localization using colorization task
}

\author{Mi Tian$^{+}$, Qiong Nie$^{+}$, Hao Shen$^{*}$, Xiahua Xia
\thanks{$^{+}$ indicates equal contributions. * indicates corresponding author. All authors are with Meituan group, Beijing, China. Email: $\{$tianmi02, nieqiong, shenhao04, xiahuaxia$\}$@meituan.com.}
}%

\begin{document}

\maketitle
\thispagestyle{empty}
\pagestyle{empty}

\begin{abstract}

Visual localization is one of the most important components for robotics and autonomous driving. Recently, inspiring results have been shown with CNN-based methods which provide a direct formulation to end-to-end regress 6-DoF absolute pose. Additional information like geometric or semantic constraints is generally introduced to improve performance. Especially, the latter can aggregate high-level semantic information into localization task, but it usually requires enormous manual annotations. To this end, we propose a novel auxiliary learning strategy for camera localization by introducing scene-specific high-level semantics from self-supervised representation learning task. Viewed as a powerful proxy task, image colorization task is chosen as complementary task that outputs pixel-wise color version of grayscale photograph without extra annotations. In our work, feature representations from colorization network are embedded into localization network by design to produce discriminative features for pose regression. Meanwhile an attention mechanism is introduced for the benefit of localization performance. Extensive experiments show that our model significantly improve localization accuracy over state-of-the-arts on both indoor and outdoor datasets.

\end{abstract}

\begin{figure*}[t]
	\centering
	\includegraphics[height=5cm]{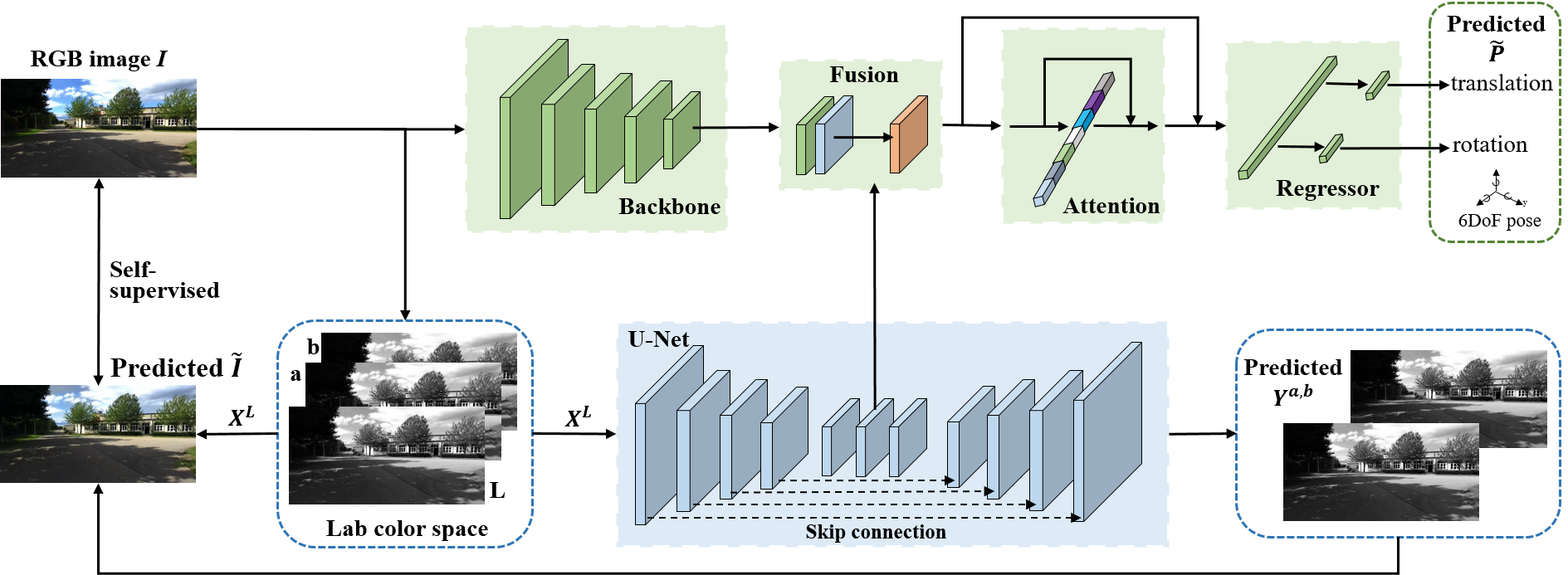}
	\caption{Schematic representation of our proposed deep auxiliary learning architecture for visual localization. Green stream is localization sub-network with a given RGB image and predicts 6DoF camera pose. Blue stream is colorization sub-network with corresponding grayscale photograph mapped from a RGB image and predicts pixel-wise color version. The feature representations of C-net are embedded into L-net.}
	\label{fig:example}
\end{figure*}
\section{INTRODUCTION}

Visual localization is the problem of estimating the camera pose of a given image related to a visual representation of a known scene. It is one of the most critical prerequisites for computer vision applications in robotics and autonomous driving. It provides fundamental support for other modules, e.g., perception and path planning by sensing where it is in the map. 

Traditional localization methods estimate camera pose from 2D-3D matching between query images and a map by applying an Perspective-n-Point (PnP) in RANSAC loop. The map is predefined with scene information that is generally presented as sparse key points with 3D geometry information and feature descriptors. The early researches aim to improve the efficiency and robustness of such 2D-3D matching by using an intermediate step of keyframe matching \cite{sattler2016efficient,sattler2012image}  or hashing algorithms \cite{andoni2015practical}. Since these methods rely heavily on local features descriptors, which are sensitive to illumination and weather changes, some literatures propose to train more robust local features and descriptors such as \cite{DeTone_2018_CVPR_Workshops} and \cite{sarlin2019coarse}. In addition, high-level semantic information is also taken into account to score the matching of images and features by the semantic consistency for visual localization \cite{shi2019visual,toft2018semantic}. Aforementioned methods are based on the traditional localization framework and convolutional neural networks (CNNs) are only used for learning semantics. The first end-to-end CNNs is proposed by \cite{kendall2015posenet} for localization. Different from traditional methods, CNN-based methods map the scene through training the network and predict corresponding 6-DoF camera pose of the given image at inference process. Thus, CNN-based methods can leverage execellent feature representation capability of deep learning for localization. Besides, using a network to map the scene makes these methods more scalable and memory-efficient than geometric methods that usually require a large database of landmarks. The main weakness is that the localization accuray is not comparable to prior geometric methods. Then many efforts are made to improve localization performance by taking advantages of more complex architectures \cite{clark2017vidloc}, geometry-aware constraints \cite{brahmbhatt2018geometry} and semantic information \cite{radwan2018vlocnet++}, etc. 

In this paper, we focus on investigating the impact of semantics on localization accuracy. State-of-the-art semantic segmentation methods incur heavy expense to collect enormous semantic annotations. Therefore we introduce a complementary self-supervised colorization task for the auxiliary learning of high-level semantic feature representations instead of using supervised semantic segmentation methods. The purpose of colorization is to convert grayscale photos to color. Colorization networks are performed in a self-supervised way as each image can be split into its intensity and its color, and the intensity can be used to predict its color. Besides, it is an intuitive sense that colorization system should be able to interpret the semantic composition of the scene (what is in the image: plants, sky, buildings, ...) as well as localize objects (where things are) before assigning a probable color to it. Therefore, it is reasonable that semantic information is hidden in the feature representations which are trained for the purpose of colorization. In fact representation learning via colorization has been studied by several literatures \cite{cao2017unsupervised,Deshpande_2017_CVPR,larsson2016learning,zhang2016colorful}. \cite{larsson2017colorization} even viewed it as a proxy task for self-supervised respresentation learning and generalized to other visual tasks well such as classification and segmentation. We build on these successes and aggregate the colorization task into our localization framework to leverage its high-level semantics and sel-supervised strategies. Moreover, an attention mechanism is introduced to boost network performance both in terms of localization results and training convergence by activating useful regions in features.

In summary, there are mainly three contributions in this paper: 
\begin{itemize}
\item We propose a novel deep auxiliary learning architecture that includes a localization sub-network and a complementary colorization sub-network. As our knowledge, it is the first time that automatic colorization is employed to help localization task by embedding high-level semantic representations. 
\item The effect of attention mechanism on localization is discussed and a localization-specific attention strategy is designed to selectively activate salient objects and useful regions for localization. 
\item We present extensive experimental evaluations on both indoor and outdoor datasets by comparing our approach with state-of-the-art methods. The results show that our method achieve excellent localization ability compared with other approaches. 
\end{itemize}

The remainder of this paper is structured as follows: Sec. 2 gives a brief review about CNN-based localization, automatic colorization and attention strategies etc. The proposed architecture is presented in Sec. 3. Sec. 4 extensively evaluates our approach comparing with current state-of-the-arts. Finally, conclusions are drawn in Sec. 5.

\section{Related Work}
\textbf{CNN-based Localization}. Unlike mapping processing of traditional methods aimed to structure a lot of feature points with geometry information and descriptor, and the more points, the better localization, CNN-based methods mapped scene information by learning weights of network and were more capable of handling large scale scenarios. Thus, many CNN-based methods for visual localization were explored and exploited in prior works. PoseNet \cite{kendall2015posenet} was the first proposed approach which utilized base architecture of GoogLeNet to directly regress 6DoF camera pose with a given RGB image. Following closely, Kendal et at. \cite{kendall2016modelling} utilized Bayesian CNNs to estimate the uncertainty of predicted pose. Melekhov et al. \cite{melekhov2017image} proposed a symmetric encoder-decoder architecture. Walch et al. \cite{walch2017image} and Clark et al. \cite{clark2017vidloc} introduced Long-Short Term Memory (LSTM) to exploit the advantage of feature learning from constraint of temporal smoothness of the video stream. Kendall et al. \cite{kendall2017geometric} and Brahmbhatt  \cite{brahmbhatt2018geometry} proposed geometry-aware constraints as extra loss terms to boost training. Brachmann et al.  Valada et al. \cite{valada2018deep}, Lin et al. \cite{lin2019deep} and Xue et al. \cite{xue2019beyond} joint learned visual odometry as an auxiliary to improve localization that were benefit from relative motion consistency of sequence images to penalize these contradictive pose prediction. Different from above works, Radwan et al. \cite{radwan2018vlocnet++} firstly introduced multitask learning framework for visual localization, odometry estimate and semantic segmentation. Through introducing semantics from segmentation and exploiting the mutual benefit of each task, results of localization were more significantly improved than other methods. However, it should be noted that costly manual annotations and multi-sensor information are required that violates the advantages of CNN-based methods in large scale scenarios meanwhile limits generalization of the algorithm. Beside above methods predicting pose directly, Brachmann et al. \cite{brachmann2017dsac,brachmann2018learning} and Zhou et al \cite{zhou2020kfnet} predicted the scene 3D coordinates for pixels by training CNNs with ground truth scene coordinate and then calculated 6DoF pose by PnP with RANSAC.  Laskar et at \cite{laskar2017camera} and Ding et al \cite{ding2019camnet} found the nearest images as key frames from database using retrieval model and regressed relative pose between query image and key frames using relative pose regression model.  In this work, we aim to retain the edge of CNN-based methods and benefit from aggregating scene-specific high-level semantics into localization. To end this, we propose an auxiliary learning using self-supervised colorization to produce useful semantic feature representations

\textbf{Colorization} is a boutique computer graphics task that aims to recover a plausible color version for a given grayscale photograph. For the purpose of auxiliary learning, we mostly focus on CNN-based methods of colorization. Zhang et al. \cite{zhang2016colorful} implemented colorization using classification task to increase the diversity of results. Larsson et al. \cite{larsson2016learning,larsson2017colorization} exploited both low-level and semantic representations to predict per-pixel color histograms. Isola et al. \cite{isola2017image} and Cao et al. \cite{cao2017unsupervised} introduced conditional GANs. Deshpande \cite{Deshpande_2017_CVPR} exploited multi-modal colorization. In these works, colorization was viewed as an promising avenue for self-supervised visual representation learning and generalized well to other visual task such as objection classification, dection and segmentation that have been proven to be surprisingly useful. Considering two advantages that availability on capturing high-level visual representations to incorporate semantic parsing for scene understanding and freedom in training data to enable the capability of handling large scale scenarios, colorization task can serve visual localization perfectly. So, it is reasonable that using self-supervised colorization to auxiliary learn localization can overcome the drawback of prior CNN-based methods and be a breakthrough exploration in visual localization.

\textbf{Attention mechanism} is widely applied in CNN-based algorithms due to general performance improvements in variety of tasks, from object detection \cite{wang2017residual,liu2018decidenet}, semantic segmentation \cite{harley2017segmentation,chen2016attention} to image captioning \cite{anderson2018bottom,li2017image}. The method proposed by \cite{hu2018squeeze} was a typically implementation by modeling channel-wise relationships in a computationally efficient manner and designed to enhance the representational power of basic modules throughout the network. The other work \cite{yang2018weakly} focused on finding salient objects in images for sentiment classification by generating activation feature map. Considering when human localize where they are by visual perception, reference landmarks in the scene are more dependent. It suggests that enhancing feature representations of regions of interest is resonable, rather than that all the feature representations contribute equally for  localization. In our work, following these methods, we introduce and discuss localization-specific attention mechanism to either optimize channel-wise information or activate regions of interest for the purpose of improving localization performance.

\section{Approach}
Our proposed deep auxiliary learning architecture is depicted in Figure 1. It consists of two sub-nets, namely localization sub-network (L-net) and colorization sub-network (C-net). Given a RGB image and corresponding grayscale image for L-net and C-net respectively, absolute camera pose and color version are jointly predicted. As a complementary task, C-net is trained to learn colorization, but the goal is to provide scene-specific feature representations to help pose regression by aggregating high-level semantic into L-net. Beyond that, we design an attention mechanism by selectively activating meaningful regions for localization task. In the following, L-net and C-net as well as this localization-specific attention strategy will be presented in detail.
\subsection{Colorization Sub-Network}
Following the typical CNN-based colorization algorithm \cite{zhang2016colorful}, we perform our task in CIE Lab color space as distances in this space model perceptual distance. Our C-net predicts the corresponding a and b color channels $\tilde{Y}^{a,b}$ (we denote ground truth as $Y^{a,b}$) with lightness L channel $X^L$ as input. Different from \cite{zhang2016colorful} using a VGG-style architecture, we adopt the lightweight U-Net \cite{ronneberger2015u} that has four downsampling and followed four upsampling operations as well as skip connections employed between layers with same resolution. In order to search for a prediction of vibrant and realistic colorization to increase the diversity of colors, some algorithms think colorization as a classification problem and use cross-entropy loss. In our case, colorization is designed as an complementary task for auxiliary learning of localization, we do not overly entangle with colorization performance but focus on localization improvement by aggregating semantic representation. Thus, we define a $l_1$ Euclidean distance loss function between predicted and ground truth color version to constrain the weights updating of C-net.  
$$L_c(X^L)=\sum_{h,w} {\lVert {\tilde{Y}^{a,b}-Y^{a,b}}\rVert}_1$$

Some previous works evaluated how representation learning via colorization can serve for classification task. It was demonstrated that best classification performances are achieved when feature representations from the maximum downsampled layer are contributed into linear classifiers. From our C-net, we follow this conclusion and  choose 16-times downsampled feature representations with respect to the input dimension in order to introduce high-level semantic information into localization. We define these representation as $M_c$ for the ease of following description. 
\subsection{Localization Sub-Network}

Given a RGB image $I\in\mathbb{R} ^{H \times W \times 3}$, L-net can predict the absolute pose $P=[x, q]$, where $x\in\mathbb{R} ^{3}$ denotes the translation and $q\in\mathbb{R} ^{4}$ denotes the rotation in a unit quaternion representation. L-net consists of four parts, a backbone for feature extraction, a fusion operation, an attention module and a regressor to predict associated pose. 

\textbf{Backbone} Following typical localization architecture, our backbone is designed with five standard residual blocks which have the same bottleneck structure and unit setting with ResNet-50 \cite{he2016deep} architecture. The output feature representations of the Res5 block is 16-times downsampled with respect to input image, and we define it as $M_l$.      

\textbf{Fusion operation} The aim of this step is to fuse representations $M_l$ from localization backbone and semantic representations $M_c$ from colorization. Since L-net and C-net are two independent tasks with different input modalities and prediction purposes, it is crucial to intelligently select helpful information from $M_c$ and avoid introducing irrelevant feature representations which may have a negative effect on pose regression. To this end, we apply a typical linear weighting and activating operations after the cross-channel concatenation between $M_l$ and $M_c$ to active elements that are useful for pose regression. This fusion operation can be formulated as:  
$$M_{fuse}=max(W \ast (M_l \oplus M_c)+b,0 )$$

Where  $\oplus$ represents concatenation across channels; $W$ and $b$ are weight parameters updated by learning. $M_{fuse}$ is the feature representation after fusion. This fusion operation is completed by a convolution and activate layer in our experiments.

\textbf{Attention module}. As previously analyzed, different feature representations do not equally contribute to localization task. Thus, an attention module is designed upon prior similar works \cite{hu2018squeeze} and \cite{yang2018weakly}. Reference to human visual localization mechanism, regions of interest consisting of salient objects or special texture ought to be enhanced in localization-specific attention module.

This attention operation can be described with two steps. The first squeeze and excitation process. Activating $M_{fuse}$ with an channel-wise vector $V=GMP(M_{fuse})$, where $V\in\mathbb{R}^C$ obtained by global max pooling operation of squeezing spatial information into a global representation. And excitation is completed by per-channel multiplication between $M_{fuse}$ and $V$.
$$M_{atten}=M_{fuse}\otimes (GAP(M_{SaE}))$$
$$M_{SaE}=M_{fuse}\otimes GMP(M_{fuse})$$

The next step aims to achieve an attention mask to weight different regions of interest in feature representations $M_{fuse}$. Such mask is computed by applying a cross channel global average pooling operation on $M_{SaE}$ and is a single channel map with high response to regions of interest. As shown in Figure 2, some regions with salient objects, e.g., gates and windows of building, bicycles and stairs are activated to impel network to learn localization from more useful feature respresentations. Finaly, to not loss information, we fuse the original holistic representations $M_{fuse}$ with the region enhanced representations $M_{atten}$ via aforementioned fusion operation and use them for the final regression.

\textbf{Regressor}. After the attention module, pose regressor is designed with three inner fully connected layers with 2048 neurons, 3 neurons and 4 neurons respectively. The first one embeds feature representations into a high-dimensional vector and the last two separately output absolute position and orientation. In order to tackle the over parameterization problem of 4-dimension unit quaternion, we follow the method proposed by \cite{brahmbhatt2018geometry} and adopt a de-parameterization strategy during training process. The idea is to convert 4-dimentional unit quaternion $q=[u,\textbf{v}]$ ($u$ is a scalar and $\textbf{v}$ is a 3 dimensional vector) into 3 dimensions by the following formula:
$$\log{q}=\frac{\textbf{v}}{\Vert\textbf{v}\Vert}\cos{u^{-1}}$$

Finally, we define the loss function with typical $l_2$ Euclidean distance to constrain the weights updating of L-net. Meanwhile a hyper parameter is introduced to balance translation and rotation error.

$$L_l(I)= {\lVert {\tilde{x}-x}\rVert}_2+\beta_{intra}\ast{\lVert {\tilde{q}-q}\rVert}_2$$

\begin{figure}[t]
	\centering
	\includegraphics[height=2.4cm]{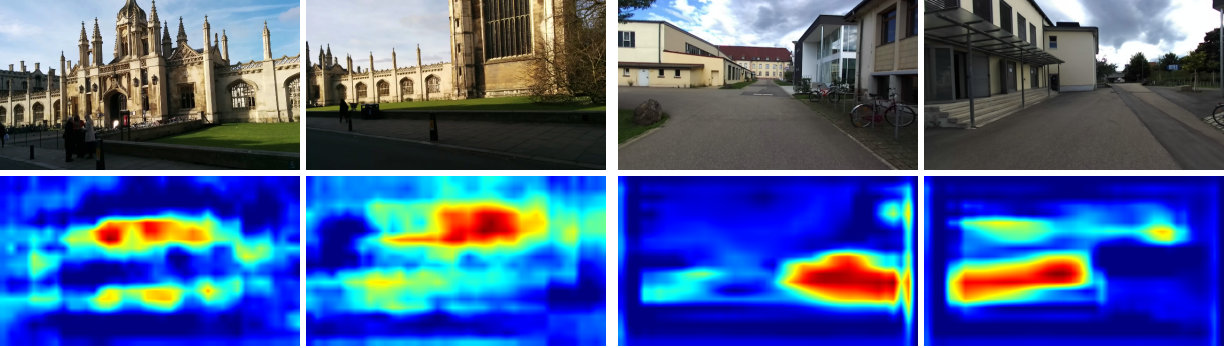}
	\caption{Visualization of attention masks from outdoor datasets. The first row denotes raw images, and the second row denotes corresponding attention masks from attention module. For each attention mask subfigure, red regions denote high response and blue regions denote low response.}
	\label{fig:example1}
\end{figure}
\subsection{Joint Learning} 
Our proposed architecture can be viewed from the perspective of multi-task learning as a soft-parameter sharing approach. The objective of auxiliary colorization task is to enable the model to learning representations that are helpful for the main localization task. Thus we adopt a joint training strategy that allows the model to learn beneficial representations and also to be more robust against random data noise. The final loss function is then defined as:
$$L=\beta_{inter} \ast L_c(X^L)+ L_l(I)$$

The former term corresponds to the colorization regularizer and the latter describes localization related loss function. Similar to aforementioned intra-task parameter, $\beta_{inter}$ is also a hyper parameter that keeps balancing inter-task loss terms. It can be trained online or preset.

\section{Experiments}
In this section, we evaluate our visual localization method compared with state-of-the-art methods both on indoor and outdoor datasets. The experimental results demonstrate a surprising performance improvement of our auxiliary learning method. 

\begin{table*}[]
\caption{Comparison of median localization error with prior CNN-based models on outdoor datasets}
\begin{tabular}{lccccc|c|c}
\hline\noalign{\smallskip}
               & \multicolumn{4}{c}{Cambridge landmarks}                  &            & Oxford robotcar &            \\
Method         & King's College & Shop Facade & Church     & Old Hospital & Average    & Loop            & DeepLoc    \\ 
\hline\noalign{\smallskip}
PoseNet15 \cite{kendall2015posenet}                & 1.66m, $4.86^\circ$     & 1.41m, $7.81^\circ$  & 2.45m, $7.96^\circ$ & 2.62m, $4.90^\circ$   & 2.04m, $6.23^\circ$ & 20.29m, $17.45^\circ$    & 2.42m, $3.66^\circ$ \\
PoseNet16  \cite{kendall2016modelling}            & 1.74m, $4.06^\circ$     & 1.25m, $7.54^\circ$  & 2.11m, $8.38^\circ$ & 2.57m, $5.14^\circ$   & 1.92m, $6.28^\circ$ & -               & 2.24m, $4.31^\circ$ \\
SVS-Pose \cite{naseer2017deep}                      & 1.06m, $2.81^\circ$     & \textbf{0.63m}, $5.73^\circ$  & 2.11m, $8.11^\circ$ & 1.50m, $4.03^\circ$   & 1.33m, $5.17^\circ$ & -               & 1.61m, $3.52^\circ$ \\
PoseNet17 \cite{kendall2017geometric}            & 0.99m, $1.06^\circ$    & 1.05m, $3.97^\circ$  & 1.49m, $3.43^\circ$ & 2.17m, $2.94^\circ$   & 1.43m, $2.85^\circ$ & -               & -          \\
PoseNet17(geo) \cite{kendall2017geometric}    & 0.88m, $1.04^\circ$     & 0.88m, $3.78^\circ$  & \textbf{1.57m}, $\mathbf{3.32^\circ}$ & 3.20m, $3.29^\circ$   & 1.63m, $2.86^\circ$ & -               & -          \\
GPPoseNet \cite{cai2019hybrid}                        & 1.61m, $2.29^\circ$     & 1.14m, $5.73^\circ$  & 2.93m, $6.46^\circ$ & 2.62m, $3.89^\circ$   & 2.08m, $4.59^\circ$ & -               & -          \\
MapNet \cite{brahmbhatt2018geometry}           & 1.07m, $1.89^\circ$     & 1.49m, $4.22^\circ$  & 2.00m, $4.53^\circ$ & 1.94m, $3.91^\circ$   & 1.63m, $3.64^\circ$ & 9.84m, $\mathbf{3.96^\circ}$      & -          \\
MLFBPPose \cite{wang2019discriminative}      & 0.76m, $1.72^\circ$     & 0.75m, $5.10^\circ$  & 1.29m, $5.01^\circ$ & 1.99m, $2.85^\circ$   & 1.20m, $3.67^\circ$ & -               & -          \\
\textbf{Ours}                                                                  & \textbf{0.72m}, $\mathbf{0.55^\circ}$     & 0.73m, $\mathbf{3.46^\circ}$  & 1.65m, $3.34^\circ$ & \textbf{1.67m}, $\mathbf{1.14^\circ}$   &\textbf{1.19m}, $\mathbf{2.12^\circ}$ & \textbf{6.49m}, $5.06^\circ$      & \textbf{0.48m}, $\mathbf{2.76^\circ}$ \\ 
\hline\noalign{\smallskip}
\end{tabular}
\end{table*}

\subsection{Datasets}
We benchmark performance of our method on three outdoor datasets: DeepLoc \cite{radwan2018vlocnet++} , Cambridge landmark \cite{kendall2015posenet} and Oxford robotcar \cite{RobotCarDatasetIJRR, RCDRTKArXiv}. DeepLoc is a large-scale urban outdoor localization dataset in an area spanning $110m\times130m$. It is collected by a robot moving along a loop road in an university campus. The captured images are with resolution of $1280\times720$ and the ground truth is computed by a LiDAR-based SLAM algorithm. The other dataset King's College, with an area of $140m\times40 m$, is widely used for benchmarking localization tasks. It is collected around a landmark building and the ground truth pose is obtained by a Structure From Motion algorithm. Both datasets are very challenging since training data and test data are captured from different points in time and distinct walking paths. Moreover, significant urban clutter such as pedestrians and vehicles are present in King's College dataset, and scene views vary largely within different paths of DeepLoc.

A well-known indoor dataset - Microsoft 7-Scene \cite{shotton2013scene} is also used for evaluating our method. Seven difference scenes are recorded from a handheld Kinect RGB-D camera at $640\times480$ resolution and ground truth camera pose is provided by KinectFusion algorithm. The existence of motion blur and weak texture under office environment makes this 7-scene dataset very challenging. 

\begin{figure}[t]
	\centering
	\includegraphics[height=3.6cm]{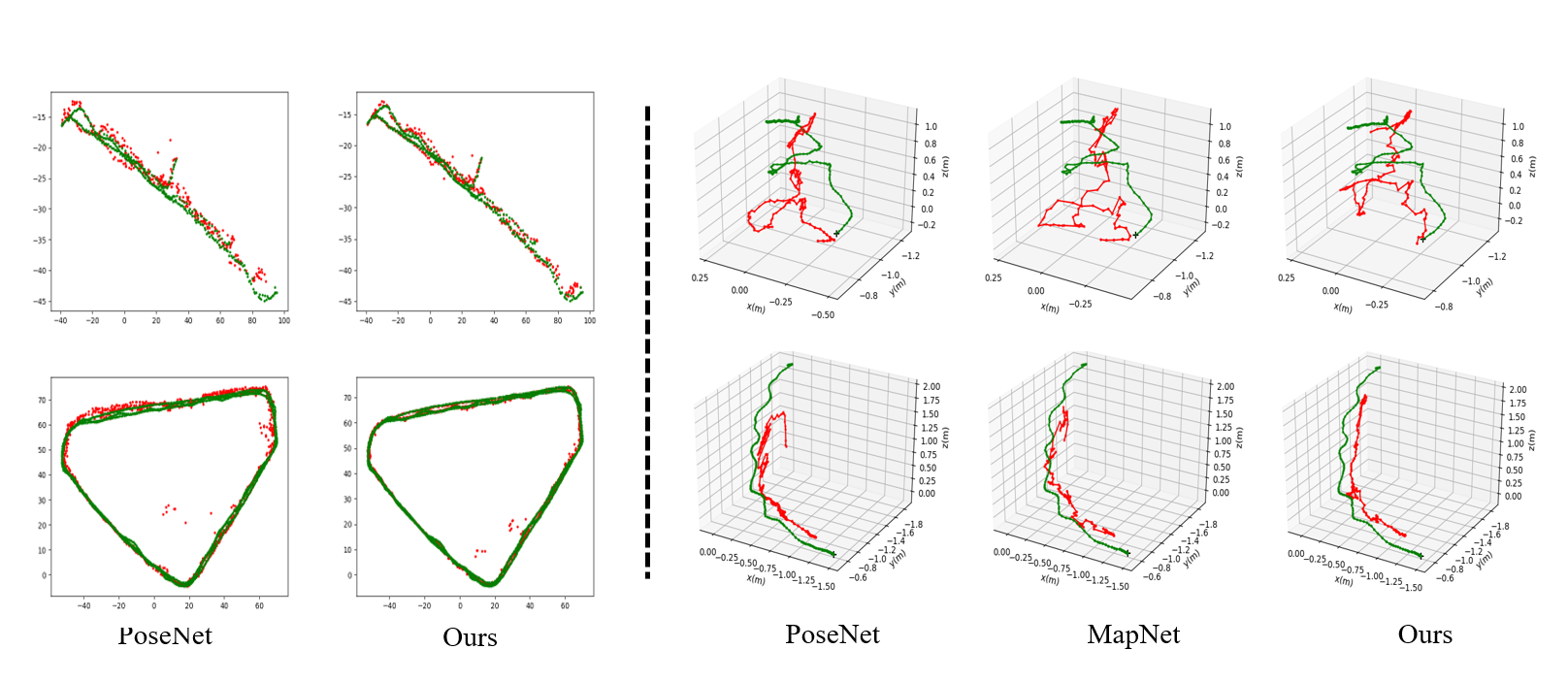}
	\caption{Visual localization results with different methods on outdoor (Left: the first row is king's College and the second row is DeepLoc) and indoor (Right: both two rows are stairs scene of Microsoft 7Scene) datasets. For each subfigure, the ground truth camera trajectory is the green lines/points, and the red lines/points show the camera pose predictions.}
	\label{fig:example2}
\end{figure}

\subsection{Implementation Details}

Although preprocessing is widely used for many visual tasks, prior works demonstrate that typical preprocessing techniques such as cropping and mirroring can not yield performance improvements for localization. In some cases, they even negatively affect the pose accuracy. In our experiments, we only take already proved well-performanced preprocessing steps like resize input images into $320\times240$ and normalized by pixel mean subtraction and standard deviation division operation.

We use the Adam solver for optimization with $\beta_1=0.9$, $\beta_2=0.99$ and $\xi=10^{-10}$. We initialize five residual blocks of L-net with weights of ResNet-50 pre-trained on ImageNet and remaining layers with Gaussian distribution. The balance parameters, $\beta_{intra}$  and $\beta_{inter}$ are set as 3 and 0.2 for all datasets. Then we train all layers with mini-batch size of 10. The learning rate of the backbone layers of L-net and all layers of C-net are initialized as 0.0003. All other layers of L-net are initialized as a learning rate of 0.001, and both learning rates decay by a power=0.9 every 10 epochs. The total number of iteration is about 150 epochs for two outdoor datasets and 80 epochs for indoor Microsoft 7Scene dataset. The work is implemented based on Tensorflow deep learning library. All the experiments are performed on a NVIDIA Titan V GPU with 16GB on-board memory.

\subsection{Evaluation on Outdoor Datasets}
The evaluation results of our method against prior CNN-based methods are shown in Table 1. PoseNet \cite{kendall2015posenet} and its variant BaysianPoseNet \cite{kendall2016modelling} as well as ours have input of a single image. SVS-Pose \cite{naseer2017deep} leverages additional depth information to do data augmentation in 3D space. VLocNet \cite{valada2018deep} takes advantages of sequential information. And VLocNet++ \cite{radwan2018vlocnet++}, as a multi-task framework for odometry, localization and semantic segmentation learning, requires much more information such as successive images, semantic segmentations annotation and depth information. Results of PoseNet2 \cite{kendall2017geometric} on DeepLoc and VLocNet++ \cite{radwan2018vlocnet++} on King's College are absent since no publish data is available. 

According to the median translation and rotation error indicated in Table 1, our method outperforms all other algorithms on King's College. `` - ’' denotes no data provided. On Deeploc dataset, our method also shows outstanding localization performance. Although VLocNet++ has slightly smaller localization error than ours, it should be noted that our method requires a single image for both training and inference, while VLocNet++ should be trained with tremendous semantic segmentation annotation cost and ground truth depth. Also, from camera pose trajactories for test sequences from DeepLoc and King's College datasets shown in Figure 3, predicted results with our method have relatively few outliers and higher localization accuracy.

\subsection{Colorization Results}
Colorization results from our method are presented in Figure 4. Four groups of image colorization results are shown. Three each group, the first image is the original color image, the second image is grayscale and the last one is the colorization image. Apparently our colorization results are quite natural and indistinguishable from original image. In order to quantitatively evaluate colorization performance, we display statistical analysis in Table 2 by counting the percentage of pixel whose colorized value is close to the corresponding pixel in the original image. We define this closeness by judging if the color difference is smaller than a threshold (eg. 5 or 10). The results show that our colorization version is much similar to the original color. The main differences exist in some areas with bright and unfixing colors like sky, as the regression strategy used in our colorization has an averaging effect on color prediction.    
\begin{figure}[t]
	\centering
	\includegraphics[height=4cm]{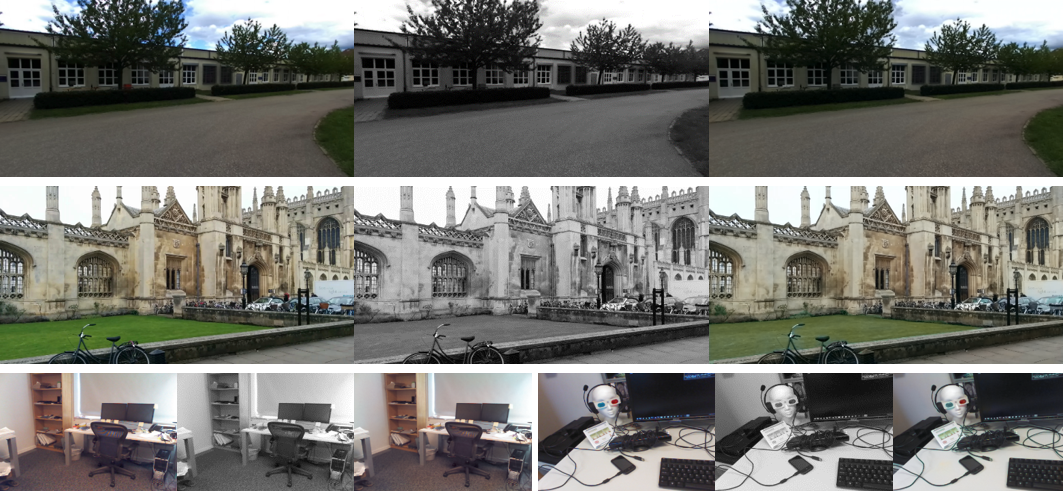}
	\caption{Colorization results on outdoor and indoor datasets. From top to buttom, the three scenes are from DeepLoc, King's College, and Microsoft 7Scene datasets. Every three images are a group, and from left to right are raw RGB images, corresponding grayscale L channel images and predicted color images respectively.}
	\label{fig:example3}
\end{figure}

\begin{table}
	\setlength{\tabcolsep}{10pt}
	\scriptsize
	\begin{center}
		\newcommand{\tabincell}[2]{\begin{tabular}{@{}#1@{}}#2\end{tabular}}
		\caption{Colorization results on outdoor datasets}
		\label{table:headings1}
		\begin{tabular}{ccccc}
			\hline\noalign{\smallskip}
			Scene                & \multicolumn{2}{c}{DeepLoc}    & \multicolumn{2}{c}{King's College}        \smallskip   \\
			\hline\noalign{\smallskip}
			Threshold   & @5          & \multicolumn{1}{l |}{@10} & @5         & @10                 \smallskip \\ 
			\hline\noalign{\smallskip}
			Accuracy            & 94.04$\%$     & \multicolumn{1}{l|}{97.84$\%$}   &93.42$\%$    & 95.87$\%$           \smallskip         \\
			\hline\noalign{\smallskip}
		\end{tabular}
	\end{center}
\end{table}

\subsection{Ablation Studies}
Since auxiliary learning and attention strategy are both introduced in our method, in this section we will separately analyze their influences on localization performance. Therefore we design three different network architectures for different ablative levels. At first, we evaluate the localization performance of the baseline L-net by cutting the representation input from C-Net and removing attention module. In the second network, we re-add C-Net for auxiliary learning to L-net but keep removing attention module. The third network is our whole architecture with both auxiliary learning and attention mechanism. 

These three networks are trained with the same initialization and hyper-parameter setting, and tested on two outdoor datesets. The results are shown in Table 3. Apparently our introduced auxiliary learning strategy via colorization leads to significant performance improvements. This improvement is more evident from the test on DeepLoc. We believe that this is because environment situations in DeepLoc are more complicate than those in King's College. Images from King's College is mainly based on a landmark building. But DeepLoc is a normal urban scenario with various streets and buildings. Semantic information from this situation seems more difficult to be learned by localization network directly. In this case, semantic feature representations learned via colorization are especially beneficial for localization. Moreover, attention mechanism also contributes to localization task by proving accuracy improvement as well as training convergence acceleration. For instance, during the test in Deeploc, required training epoch for the whole network is 155, but it increases to 189 when we remove the attention module.    

\begin{table}[]
\begin{center}
\caption{Comparison of median translation error (m) and rotation error ($^\circ$) with different ablative levels on outdoor datasets}
\begin{tabular}{cccc}
\hline\noalign{\smallskip}
Scene          & \begin{tabular}[c]{@{}c@{}}Auxiliary\\ learning\end{tabular} & \begin{tabular}[c]{@{}c@{}}Attention\\ module\end{tabular} & Median error \\ 
\hline\noalign{\smallskip}
               &                                                            &                                                           & 0.67m, $3.51^\circ$   \\

DeepLoc        &   \checkmark                                                        &                                                          & 0.50m, $2.88^\circ$   \\
               &  \checkmark                                                        &  \checkmark                                                        & \textbf{0.48m}, $\mathbf{2.76^\circ}$   \\ 
\hline\noalign{\smallskip}
               &                                                             &                                                         & 0.86m, $1.61^\circ$   \\

King's College &  \checkmark                                                           &                                                        & 0.78m, $0.90^\circ$  \\
               &  \checkmark                                                            & \checkmark                                   & \textbf{0.72m}, $\mathbf{0.55^\circ}$   \\ \hline
\end{tabular}
\end{center}
\end{table}

\subsection{Comparison with Joint Semantic Segmentation}
In this section we replace our colorization task by semantic segmentation and compare their auxiliary learning function on localization. This means that we keep C-net unchanged but its training purpose turns into semantic segmentation. Thus the input of C-net should be changed to a RGB image. And the output becomes into a pixel-wise segmentation prediction whose channel number is equal to the number of semantic  annotation categories. This new network is then trained with DeepLoc dataset. Since semantic segmentation labels are available for only half training data in this dataset, we train the whole network (semantic segmenation task and localization task) with semantic segmentation labeled images. Then we fine tune L-net with rest semantic unlabeled images by fixing weights of C-net. The localization results after both stage are presented in Table 4.

According to the results shown in Table 4, it seems that colorization and semantic segmentation equally contribute when they are considered as auxiliary task for localization. But semantic segmentation suffers from additional annotation cost. 

\begin{table}
	\setlength{\tabcolsep}{4pt}
	\begin{center}
		\scriptsize
		\newcommand{\tabincell}[2]{\begin{tabular}{@{}#1@{}}#2\end{tabular}}
		\caption{Comparison of median translation error (m) and rotation error ($^\circ$) with joint semantic segmentation method on DeepLoc dataset}
		\begin{tabular}{cccc}
			\hline\noalign{\smallskip}
			\multicolumn{1}{l}{} & \multicolumn{1}{c}{Stage-1} & \multicolumn{1}{c|}{Stage-2} & \multicolumn{1}{c}{Ours} \smallskip \\
			\hline\noalign{\smallskip}
			Median error                   & 0.64, $3.18^\circ $                 & \multicolumn{1}{c|}{0.54, $2.70^\circ$}                  & \textbf{0.48}, $\mathbf{2.76^\circ}$      \smallskip\\
			\hline
		\end{tabular}
	\end{center}
\end{table}
Above evaluations are based on localization performance and segmentation is considered as auxiliary task. Here we present semantic segmentation performance and compare it with other segmentation algorithms. The experiment is performed on DeepLoc dataset and the Intersection over Union (IoU) score for major individual categories as well as the mean IoU are shown in Table 5. The segmentation images can be seen in Figure 5. From the results, our method outperforms AdaptNet \cite{valada2017adapnet} and DeepLabv3+ \cite{chen2018encoder} as well as is very close to VLocNet++ that is attributed to the warping computation by successive images and depth information. 
\begin{table}
	\setlength{\tabcolsep}{3pt}
	\begin{center}
		\scriptsize
		\caption{Comparison of semantic segmentation prediction with state-of-the-art methods on DeepLoc dataset}
		\begin{tabular}{lccccccc}
			\hline\noalign{\smallskip}
			Approach   & Sky   & Road  & Sidewalk & Grass & Vegetation & Building & MIoU \smallskip \\ 
			\hline\noalign{\smallskip}
			AdapNet \cite{valada2017adapnet}   & 94.65 & 98.98 & 64.97    & 82.14 & 84.48      & 87.68    & 78.59 \\
			DeepLabv3+ \cite{chen2018encoder} & 94.26 & 98.46 & 81.60    & 90.94 & 91.07      & 94.20    & 78.30 \\
			VLocNet++ \cite{radwan2018vlocnet++} & \textbf{95.84} & \textbf{98.99} & 80.85    & 88.15 & 91.28      & \textbf{94.72}    & \textbf{80.44} \\ 
			Ours       & 95.49 & 98.42 & \textbf{81.30}    & \textbf{91.26} & \textbf{91.55}      & 94.60    & 79.45 \\ 
			\hline\noalign{\smallskip}
		\end{tabular}
	\end{center}
\end{table}

\begin{figure}[t]
	\centering
	\includegraphics[height=3.2cm]{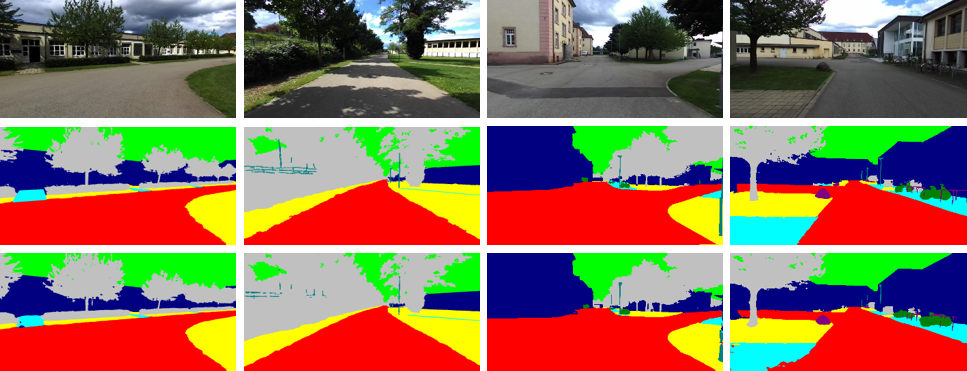}
	\caption{Semantic segmentation prediction with our proposed architecture on outdoor DeepLoc dataset. From top to bottom, the three rows denote raw RGB images, ground truth, segmentation prediction. For each subfigure of segmentation prediction, major categories: red, yellow, blue, gray, green denote road, grass, building, vegetation, sky respectively.}
	\label{fig:example4}
\end{figure}

\begin{table*}[]
\begin{center}
\caption{Comparison of median translation error and rotation error  with various CNN-based methods on Microsoft 7-Scene dataset}
\begin{tabular}{lcccccccc}
\hline\noalign{\smallskip}
Method                                                                 & Chess                         & Fire                           & Heads                       & Office                       & Pumpkin                     & Kitchen                       & Stairs                       &Average    \\ 
\hline\noalign{\smallskip}
PoseNet15 \cite{kendall2015posenet}                & 0.32m, $8.12^\circ$     & 0.47m, $14.4^\circ$  & 0.29m, $12.0^\circ$ & 0.48m, $7.08^\circ$   & 0.47m, $8.42^\circ$ & 0.59m, $8.64^\circ$    & 0.47m, $13.8^\circ$  & 0.44m, $10.8^\circ$  \\
PoseNet16  \cite{kendall2016modelling}            & 0.37m, $7.24^\circ$     & 0.43m, $13.7^\circ$  & 0.31m, $12.0^\circ$ & 0.48m, $8.04^\circ$   & 0.61m, $7.08^\circ$ & 0.58m, $7.54^\circ$    & 0.48m, $13.1^\circ$  & 0.47m, $9.81^\circ$  \\
PoseNet17 \cite{kendall2017geometric}            & 0.14m, $4.50^\circ$     & 0.27m, $11.8^\circ$  & 0.18m, $12.1^\circ$ & 0.20m, $5.77^\circ$   & 0.25m, $4.82^\circ$ & 0.24m, $5.52^\circ$    & 0.37m, $10.6^\circ$  & 0.24m, $7.87^\circ$   \\
PoseNet17(geo) \cite{kendall2017geometric}    & 0.13m, $4.48^\circ$     & 0.27m, $11.3^\circ$  & 0.17m, $13.0^\circ$ & 0.19m, $5.55^\circ$   & 0.26m, $4.75^\circ$ & 0.23m, $5.35^\circ$    & 0.35m, $12.4^\circ$  & 0.23m, $8.12^\circ$   \\
GPPoseNet \cite{cai2019hybrid}                       & 0.20m, $7.11^\circ$     & 0.38m, $12.3^\circ$  & 0.21m, $13.8^\circ$ & 0.28m, $8.83^\circ$   & 0.37m, $6.94^\circ$ & 0.35m, $8.15^\circ$    & 0.37m, $12.5^\circ$  & 0.31m, $9.95^\circ$  \\
MapNet \cite{brahmbhatt2018geometry}           & 0.08m, $\mathbf{3.25^\circ}$     & 0.27m, $11.7^\circ$  & 0.18m, $13.3^\circ$ & 0.17m, $\mathbf{5.15^\circ}$   & 0.22m, $4.02^\circ$ & 0.23m, $\mathbf{4.93^\circ}$    & 0.30m, $12.1^\circ$  & 0.21m, $7.77^\circ$   \\
ANNet \cite{bui2019adversarial}                       & 0.12m, $4.30^\circ$     & 0.27m, $11.6^\circ$  & 0.16m, $12.4^\circ$ & 0.19m, $6.80^\circ$   & 0.21m, $5.20^\circ$ & 0.25m, $6.00^\circ$    & 0.28m, $8.40^\circ$  & 0.21m, $7.90^\circ$ \\
MLFBPPose \cite{wang2019discriminative}      & 0.12m, $5.82^\circ$     & 0.26m, $12.0^\circ$  & 0.14m, $13.5^\circ$ & 0.18m, $8.24^\circ$   & 0.21m, $7.05^\circ$ & 0.22m, $8.14^\circ$    & 0.38m, $10.3^\circ$  & 0.22m, $9.29^\circ$  \\
PoseNet20 \cite{tian20203d}                            & 0.09m, $4.39^\circ$     & 0.25m, $10.8^\circ$  & 0.14m, $12.6^\circ$ & 0.17m, $6.46^\circ$   & 0.19m, $5.91^\circ$ & 0.21m, $6.71^\circ$    & 0.26m, $11.5^\circ$  & 0.19m, $8.33^\circ$ \\
\textbf{Ours}                                                     & \textbf{0.09m}, $3.77^\circ$   & \textbf{0.23m}, $\mathbf{11.3^\circ}$  & \textbf{0.14m}, $\mathbf{11.5^\circ}$ & \textbf{0.16m}, $5.61^\circ$   & \textbf{0.18m}, $\mathbf{3.80^\circ}$ &\textbf{0.20m}, $6.35^\circ$    & \textbf{0.25m}, $\mathbf{10.2^\circ}$  & \textbf{0.18m}, $\mathbf{7.52^\circ}$  \\ 
\hline\noalign{\smallskip}
\end{tabular}
\end{center}
\end{table*}

\subsection{Evaluation on Indoor Dataset}

In addition to the above outdoor datasets, experiments are also performed on indoor dataset Microsoft 7Scene in order to fully demonstrate our localization ability. We compare our method with other CNN-based methods like PoseNet \cite{kendall2015posenet}, LSTM-Pose \cite{walch2017image}, PoseNet2 \cite{kendall2017geometric}, and MapNet/MapNet+ \cite{brahmbhatt2018geometry}. The median translation and rotation errors for each method are shown in Table 6 meanwhile camera pose trajactories for test sequences of stairs scene  are presented in Figure 3. Our method has similar accuracy with MapNet+, but outperforms other methods. To be mentioned that MapNet+ is an advanced version of MapNet by fusion additional information (eg. visual odometry) to update the weights of MapNet with self-supervised learning. Therefore a correct visual odometry algorithm is required for MapNet+. While our method is more independent as only baseline information (image and corresponding ground truth pose) is required. Moreover, our model is easier to converge than others. For example, our model takes about 80 epochs iterations to converging while MapNet takes 300 epochs iterations.

To sum up, our proposed auxiliary learning for visual localization via colorization exhibits significant accuracy-efficiency balance performance without requiring extra information. All these properties makes it suitable for many applications including indoor robots and outdoor autonomous vehicles.

\section{Conclusions}
In most works, visual localization has been implemented as a independent task and performance improvement mainly comes from the optimization of network architecture and loss constraint. Some works take multi-task learning framework by introducing semantic segmentation with the goal of exploiting mutual benefit. But these semantic segmentation methods usually require tremendous manual annotations. Our proposed method aims to overcome above drawback. By taking self-supervised colorization as auxiliary task to learn high-level semantic representation for localization, our method, which requires no additional information, achieves surprising performance improvement on both outdoor and indoor datasets. Also an attention mechanism is introduced into our localization network and leads to further performance improvement on both accuracy and efficiency. 

\bibliographystyle{IEEEtran}
\bibliography{IEEEabrv, egbib}

\end{document}